\documentclass[sigconf]{acmart}
\usepackage{subcaption}
\usepackage{graphicx} 

\usepackage{algorithm}
\usepackage{algorithmic}

\usepackage{multirow}
\usepackage{float}

\usepackage{enumitem}
\AtBeginDocument{%
  }

\newcommand{\ie}{\emph{i.e., }}

\copyrightyear{2023}
\acmYear{2023}
\setcopyright{acmlicensed}
\acmConference[MM '23] {Proceedings of the 31st ACM International Conference on Multimedia}{October 29--November 3, 2023}{Ottawa, ON, Canada}
\acmBooktitle{Proceedings of the 31st ACM International Conference on Multimedia (MM '23), October 29--November 3, 2023, Ottawa, ON, Canada}
\acmPrice{15.00}
\acmISBN{979-8-4007-0108-5/23/10}
\acmDOI{10.1145/3581783.3612051}
\begin{document}

\title{General Debiasing for Multimodal Sentiment Analysis}

\author{Teng Sun}
\affiliation{%
	\institution{Shandong University}
 \country{}
}\email{stbestforever@gmail.com}

\author{Juntong Ni}
\affiliation{%
	\institution{Shandong University}
	 \city{}
  \country{}
}\email{juntongni02@gmail.com}

\author{Wenjie Wang$^*$}
\affiliation{%
	\institution{National University of Singapore}
	 \city{}
  \country{}
}\email{wenjiewang96@gmail.com}

\author{Liqiang Jing}
\affiliation{%
	\institution{Shandong University}
	 \city{}
  \country{}
}\email{jingliqiang6@gmail.com}

\author{Yinwei Wei}
\affiliation{%
	\institution{National University of Singapore}
	 \city{}
  \country{}
}\email{weiyinwei@hotmail.com}

\author{Liqiang Nie}
\affiliation{%
	\institution{\mbox{Harbin Institute of Technology (Shenzhen)}}
	\city{}
  \country{}
}\email{nieliqiang@gmail.com}
\def\authors{Teng Sun, Juntong Ni, Wenjie Wang, Liqiang Jing, Yinwei Wei, and Liqiang Nie}
\renewcommand{\shortauthors}{Sun et al.}

\thanks{*Wenjie Wang (wenjiewang96@gmail.com) is corresponding author.}

\begin{abstract}
Existing work on Multimodal Sentiment Analysis (MSA) utilizes multimodal information for prediction yet unavoidably suffers from fitting the spurious correlations between multimodal features and sentiment labels. For example, 
if most videos with a blue background have positive labels 
in a dataset, the model will rely on such correlations for prediction, while ``blue background'' is not a sentiment-related feature. To address this problem, we define a general debiasing MSA task, which aims to enhance the Out-Of-Distribution (OOD) generalization ability of MSA models by reducing their reliance on spurious correlations. To this end, we propose a general debiasing framework based on Inverse Probability Weighting (IPW), which adaptively assigns small weights to the samples with larger bias (\textit{i.e.,} the severer spurious correlations). 
The key to this debiasing framework is to estimate the bias of each sample, which is achieved by
two steps: 1) disentangling the robust features and biased features in each modality, and 2) utilizing the biased features to estimate the bias. Finally, we employ IPW to reduce the effects of large-biased samples, facilitating robust feature learning for sentiment prediction. To examine the model's generalization ability, we keep the original testing sets on two benchmarks and additionally construct multiple unimodal and multimodal OOD testing sets. The empirical results demonstrate the superior generalization ability of our proposed framework. We have released the code to facilitate the reproduction {\href{https://github.com/Teng-Sun/GEAR}{https://github.com/Teng-Sun/GEAR}}.

\end{abstract}




\maketitle

\section{Introduction}

\begin{figure}
\setlength{\abovecaptionskip}{0.1cm}
\setlength{\belowcaptionskip}{0cm}
    \centering
    \includegraphics[scale=0.55]{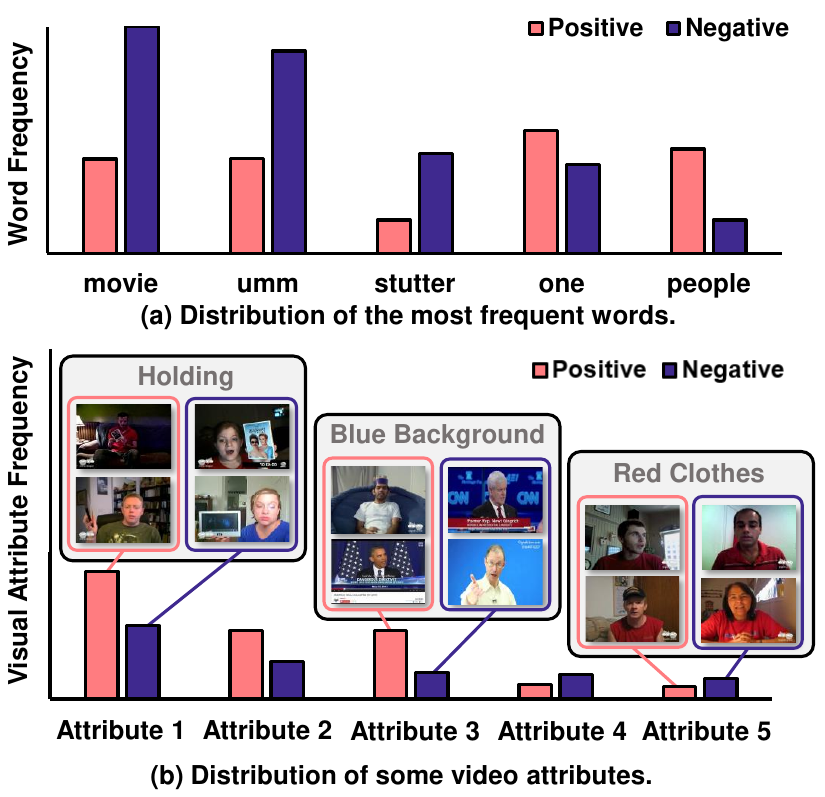}
    \caption{The distribution of the top-5 most frequent words and visual attributes.}
    \label{fig:intro}
    \vspace{-0.5cm}
\end{figure}

Sentiment analysis, a classical language understanding task, has attracted wide attention from the academy and industry. The early work chiefly utilizes users' textual reviews to analyze their sentiment~\cite{pang2004sentimental,7484319}. However, single textual modality usually has problems of polysemy and ambiguity~\cite{zadeh2016mosi, DBLP:conf/mm/ChenYYL18}. Along with the advance of social media, more and more people begin to adopt multimodal information to express their sentiments, such as video and audio~\cite{soleymani2017survey, DBLP:journals/tomccap/SunWSFN22, sun2023dual, DBLP:conf/mm/HuQFWZZX21, DBLP:conf/mm/HuFQX20}. 
The cross-modal consistency and complementarity provide rich semantic information for sentiment analysis. 
Therefore, Multimodal Sentiment Analysis (MSA) has been a popular research field in recent years~\cite{perez2013utterance, DBLP:conf/mm/YuanLXY21, DBLP:conf/mm/ZhangLZZ19, DBLP:conf/mm/YouLJY15, liu2023hs}. 

Previous studies of MSA predominantly pay attention to representation learning and multimodal fusion. For representation learning, some researchers utilize techniques like adversarial learning~\cite{mai2020modality} and multi-task learning~\cite{hazarika2020misa} to map features from different modalities into a shared representation space. Self-supervised learning~\cite{yu2021learning} is also used to incorporate unimodal information into the fusion model to aid representation learning. For multimodal fusion, previous work manages to learn cross-modal representation using sophisticated fusion mechanisms, such as tensor-based fusion~\cite{zadeh2017tensor} and graph-based fusion~\cite{zadeh2018multimodal}. In addition, some studies~\cite{rahman2020integrating,wang2022cross} also attempt to integrate modalities via pre-trained transformers (\textit{e.g.,} BERT~\cite{devlin2018bert} and XLNet~\cite{yang2019xlnet}).  

 However, existing studies usually suffer from fitting the spurious correlation between multimodal features and sentiment labels. As shown in Figure~\ref{fig:intro}, the word ``movie'' in Figure~\ref{fig:intro}(a) and the attribute ``blue background'' in Figure~\ref{fig:intro}(b) 
 show strong correlations with negative and positive sentiment labels, respectively.  
 However, ``movie'' and ``blue background'' are not reliable cues for identifying sentiment. Due to the short-cut bias~\cite{geirhos2020shortcut}, the MSA models will easily learn such spurious correlations for prediction, impairing the generalization ability in the Out-Of-Distribution (OOD) testing data, where the correlations between multimodal features and sentiment labels differ from those in the training data. For instance, Sun ~\textit{et al.}~\cite{sun2022counterfactual} pointed out the spurious correlations between textual words and sentiment labels. Nevertheless, there are also spurious correlations in video and audio modalities in addition to textual modality. 

To address the above problems, we first propose a general debiasing task for MSA, which aims to enhance the OOD generalization ability of MSA models by reducing the bad effect of multimodal spurious correlations. 
To mitigate the effect of spurious correlations, a widely used method is Inverse Probability Weighting (IPW), where a sample with strong bias will be assigned a small weight for training. To implement IPW, the key lies in estimating the bias of each sample, which depends on two steps: 1) disentangling the robust features (\textit{i.e.,} sentiment-related features such as smiling face) and biased features (\textit{i.e.,} sentiment-irrelevant features such as the blue background) in each modality, and 2) utilizing the biased features to estimate the sample bias.

To disentangle multimodal features for estimating bias weights, we propose a General dEbiAsing fRamework (GEAR) 
with three stages. 
First, we design three pairs of robust extractors and biased extractors, where each pair is used to extract the robust features and biased features in a modality. 
Second, to disentangle biased features, prior studies usually consider using Generalized Cross Entropy (GCE) loss~\cite{zhang2018generalized} to train the biased extractor and amplify the prejudice for bias estimation. 
However, such GCE loss cannot be applied to debiasing MSA since MSA is usually formulated as a regression task instead of the classification task~\cite{zadeh2017tensor}. Toward this challenge, we propose a novel Generalized Mean Absolute Error (GMAE) loss, which is specially designed to disentangle biased features in the MSA task. 
We then estimate the bias weight from biased features by calculating 
the absolute error between the outputs of the three biased extractors and sentiment labels. 
The underlying philosophy is that the biased features with strong correlations will have a lower absolute error and vice versa. 
Third, to reinforce the generalization ability, we use the estimated bias weights to adjust IPW-based Mean Absolute Error (MAE) loss for debiasing training and fuse the robust features of three modalities for prediction. 

To evaluate the generalization ability of MSA models, we construct four OOD testing sets while keeping the original testing set as an Independent and Identical Distribution (IID) testing set on two benchmarks. The empirical results demonstrate the superior generalization ability of GEAR on OOD testing sets while maintaining comparable IID performance with state-of-the-art methods. To sum up, our contributions are threefold.
\begin{itemize}[leftmargin=*]
    \item To the best of our knowledge, we are the first to formulate a general debiasing MSA task from multiple modalities. Meanwhile, to examine the generalization ability of MSA models, we contribute several multimodal OOD testing datasets. 
    
    \item We propose a novel framework GEAR, which strengthens the generalization ability of MSA models by disentangling the robust and biased features via a novel GMAE loss and estimating the bias weight of each sample for IPW-enhanced debiasing training. 
    
    \item We conduct extensive experiments on two datasets (\textit{i.e.,} MOSEI and MOSI~\cite{zadeh2016mosi}), and the experimental results demonstrate the superior generalization ability of GEAR. 
\end{itemize}

\section{Related Work}
\vspace{5pt}
$\bullet$ \textbf{Multimodal Sentiment Analysis.} In recent years, a substantial number of researchers have explored the MSA. The prior researchers mainly focused on representation learning and multimodal fusion. 
For representation learning, previous studies mainly are in three variants: 1) Shift-based models shift textual representations based on aligned nonverbal behaviors (\textit{i.e.,} audio and vision modality)~\cite{wang2019words}. 2) Shared subspace learning models map all the modalities simultaneously into modality-invariant and modality-specific representations~\cite{hazarika2020misa}. And 3) Self-supervised models generate the unimodal labels by self-supervised learning strategy and use multi-task learning to train the model~\cite{yu2021learning}. 
For multimodal fusion, according to the fusion stages, two multimodal fusion strategies are applied: 1) early fusion~\cite{zadeh2018memory,tsai2019multimodal,rahman2020integrating,yang2020cm,wang2022cross} means that the features of different modalities are combined together in an early stage. And 2) late fusion~\cite{zadeh2017tensor,liu2018efficient,dai2021multimodal} indicates that the intra-modal representation is learned first and inter-modal fusion is performed last.

Although existing studies have achieved great success, they ignore the spurious correlations between modalities and sentiment labels. Hence, Sun ~\textit{et al.}~\cite{sun2022counterfactual} is the first to settle this issue, which introduces a model-agnostic counterfactual reasoning framework (CLUE) for MSA that can leverage the positive aspects of text-based modality and mitigate potential drawbacks. However, CLUE can only handle the case of spurious correlations in a single textual modality, and cannot satisfactorily deal with multiple modalities with spurious correlations. 
To this end, we presented a general debiasing MSA network to improve the OOD generalization ability.

\vspace{5pt}
$\bullet$  \textbf{General Debiasing Methods.} The existing general debiasing methods can be divided into three categories. 
1) Debiasing with known bias types and labels. Many debiasing methods~\cite{hong2021unbiased, arjovsky2019invariant, wang2022causal} require explicit bias types and bias annotations for each training sample.
2) Debiasing with known bias types. To eliminate the costs of bias annotations, some bias-tailored studies~\cite{wang2019learning, bahng2020learning} only require bias types. 
3) Debiasing with unknown bias types. The above assumptions face limitations since manually discovering bias types strongly relies on expert knowledge and laborious labeling~\cite{nam2020learning, wang2022user}. The following work estimates the bias of each sample without knowing its bias types and labels. Nam~\textit{et al.}~\cite{nam2020learning} trained a debiased classifier from the biased classifier by utilizing GCE and relative difficulty score. Lee~\textit{et al.}~\cite{lee2021learning} learned debiased representation via disentangled feature augmentation. Fan~\textit{et al.}~\cite{fan2022debiasing} applied the methods from the previous two works to graph data debiasing.

However, multimodal data have the complex bias which is infeasible to be recognized. To this end, we resorted to the third category. Yet most existing methods are designed for image datasets and could not effectively conduct debiasing with multimodal data. Thus, we designed a debiasing framework specified for multimodal features.


\begin{figure*}
\setlength{\abovecaptionskip}{0.1cm}
\setlength{\belowcaptionskip}{0cm}
    \centering
    \includegraphics[scale=0.575]{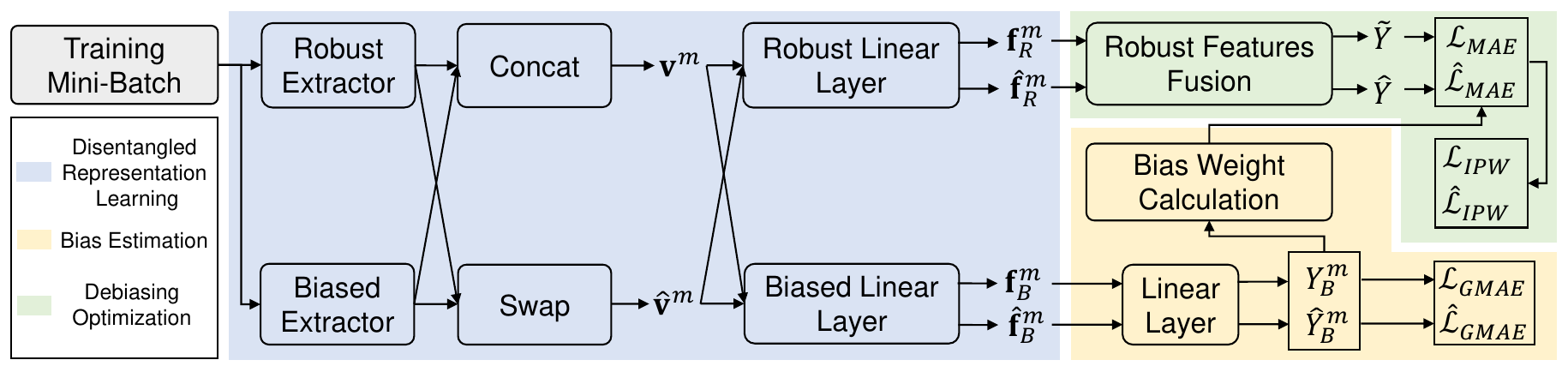}
    \caption{Illustration of the proposed general debiasing framework, which consists of disentangled representation learning,  bias estimation, and debiasing optimization.}
    \label{fig:framework}
\end{figure*}

\section{Methodology}

In this section, we first give the task formulation, and then we present the overall framework and the detailed modules.

\subsection{Task formulation}
\subsubsection{Traditional MSA Task} 
Let $\mathcal{D} = \{T_i, A_i, V_i, Y_i\}_1^N$ denote the MSA training set with $N$ training samples. Each quadruple is from a video segment, where $T_i$, $A_i$, $V_i$ and $Y_i$ denote text, audio, video, and the corresponding sentiment label of the $i$-th sample, respectively. The traditional MSA task aims to develop a model $\mathcal{F}_\theta$ which jointly utilizes three modalities  (\textit{i.e.,} $T_i$, $A_i$ and $V_i$) to predict sentiment label $Y_i$ as follows,
\begin{equation}
    \tilde{Y}_i = \mathcal{F}_\theta(T_i,A_i,V_i), 
    \label{eq:TraditionalMSATask}
\end{equation}
where $\theta$ represents the learnable parameters and $\tilde{Y}_i$ denotes the predicted label of the $i$-th sample. For clarity, we temporally omit the subscript $i$ of the training samples.

\subsubsection{General Debiasing MSA Task}

Traditional MSA models rely on complementary and consistent multimodal information for sentiment prediction. However, existing MSA models suffer from spurious correlations between multimodal features and sentimental labels. As mentioned before, the word ``movie'' is highly correlated with the negative sentiment, and the attribute ``blue background'' has a strong correlation with the positive sentiment in the MOSEI dataset. Trained with such biased data, models tend to predict samples with the word ``movie'' as negative samples and ``blue background'' as positive samples, which strongly deteriorates the generalization performance of MSA models.

To reduce the negative influence caused by multimodal spurious correlations, we formulate the general debiasing MSA task, which aims to evaluate the generalization ability of different MSA models on the OOD testing set. 
To achieve this, we propose an algorithm to automatically build OOD testing sets based on the original IID testing set. 
Specifically, the OOD testing sets have significantly different multimodal sentiment correlations from the training set. The different multimodal distributions in OOD testing sets and the training set are able to effectively evaluate whether MSA models have strong debiasing ability.


\subsection{Framework}
As can be seen in Figure~\ref{fig:framework}, the overall framework is as follows: 1) Disentangled Representation Learning Module: for a given sample with three modalities, we first disentangle the robust and biased features of each modality by robust and biased extractors, respectively. Then, we swap the robust and biased features, which synthesizes more diverse samples and facilitates the disentanglement. 2) Bias Estimation Module: we devise GMAE loss to boost the training of bias extractors. In addition, we calculate the absolute values between the prediction based on multimodal biased features and sentiment labels. The absolute values of three modalities are used to estimate the bias weight of each sample. 3) Debiasing Optimization Module: we fuse the multimodal robust features by multi-head self-attention and employ IPW-enhanced MAE loss for training robust extractors. We utilize IPW to re-weight the samples by bias weights, which discourages the influence of samples with a large bias. Each module will be elaborated on in the following sub-sections.


\subsection{Disentangled Representation Learning}
To disentangle the robust features and biased features in each modality, we simultaneously train three pairs of biased extractors and robust extractors. 
In addition, to facilitate disentanglement, we swap the robust and biased latent vectors and synthesize more diverse samples.

\subsubsection{Robust and Biased Extractors}
To extract robust and biased features in each modality, we present three pairs of the robust extractors $E_R^m$ and biased extractors $E_B^m, m\in\{t, a, v\}$.

\textbf{Extractors for Textual Modality}. In the textual modality, due to the great success of large pre-trained transformer-based language models, we utilize the pre-trained BERT as the backbone to extract textual representations of the raw text. 
Similar to the existing studies~\cite{rahman2020integrating}, we select the first [CLS] token in the last layer as the whole textual representation. After that, we employ the linear layers to map the features to the low-dimension semantic space.
We feed raw text $T$ into the textual robust extractors and biased extractors to gain robust and biased latent vectors of text (\textit{i.e., } $\textbf{v}_{\kappa}^t$). The whole structures of the textual robust and biased extractors are formulated as follows, 
\begin{equation}
\textbf{v}_{\kappa}^t = E_{\kappa}^t(T) = \textbf{W}_{\kappa}^t(BERT_{\kappa}^t(T))+\textbf{b}_{\kappa}^t, 
\end{equation}
where $\kappa\in\{R, B\}$, $\textbf{v}_{\kappa}^t\in \mathbb{R}^{d_{s}}$, $\textbf{W}_{\kappa}^t\in \mathbb{R}^{d_{s}\times d_{t}}$, $\textbf{b}_{\kappa}^t\in \mathbb{R}^{d_{s}}$, $d_{s}$ and $d_{t}$ denote the dimensions of the latent vectors and BERT's output.


\textbf{Extractors for Acoustic and Visual Modalities}. In acoustic and visual modalities, we employ the hand-crafted features extracted by Yu~\textit{et al.}~\cite{yu2021learning} from the raw data, $\textbf{A}\in \mathbb{R}^{l_a\times d_a}$ and $\textbf{V}\in \mathbb{R}^{l_v\times d_v}$. Here, $l_a$ and $l_v$ are the sequence lengths of audio and video, respectively. $d_a$ and $d_v$ are the extracted hand-crafted features dimension of audio and video, respectively. Then, we use the 1-layer Long Short-Term Memory (LSTM)~\cite{hochreiter1997long} to capture the temporal information. Similar to previous work~\cite{hazarika2020misa,yu2021learning}, we select the final states vector of LSTM as the whole modality representation. The robust and biased extractors of audio and video are similar to the ones of text except for the backbone, where we replace BERT with LSTM. The structures of the robust and biased extractors of audio and video are as follows,
\begin{equation}
\left\{
\begin{array}{l}  
\textbf{v}_{\kappa}^{a} = E_{\kappa}^{a}(\textbf{A}) = \textbf{W}_{\kappa}^{a}(LSTM_{\kappa}^{a}(\textbf{A}))+\textbf{b}_{\kappa}^{a}, \\[5pt]
\textbf{v}_{\kappa}^{v} = E_{\kappa}^{v}(\textbf{V}) = \textbf{W}_{\kappa}^{v}(LSTM_{\kappa}^{v}(\textbf{V}))+\textbf{b}_{\kappa}^{v}, 
\end{array} 
\right.
\end{equation}
where $\textbf{v}_{\kappa}^{a/v} \in \mathbb{R}^{d_{s}} $ denote the robust and biased latent vectors of audio and video, $\textbf{W}_{\kappa}^{a/v}\in \mathbb{R}^{d_{s}\times d'_{a/v}}$,  $\textbf{b}_{\kappa}^{a/v} \in \mathbb{R}^{d_{s}}$, and $d'_{a/v}$ are the dimension of the output of LSTM.


\subsubsection{Diversify Samples via Swap}
 We argue that the diversity of samples is of vital importance for disentanglement. With the visual modality as an example, the facial expressions reflect sentiment precisely. It is interesting to note that yellow can make people feel happy, so abundant samples contain the pair of (\textit{Smile}, \textit{Yellow}) which means a smile and yellow background coexist in an image. On the contrary, purple tends to make people feel sad, so ample samples contain (\textit{Frown}, \textit{Purple}). In the two pairs, the facial expressions are robust attributes and the colors of the background are biased attributes. While the above architecture disentangles the robust features and bias features, $E_R^m$ and $E_B^m$ are still mainly trained with small amounts of samples that have poor diversity. Thereby, the above architecture is able to disentangle (\textit{Smile}, \textit{Yellow}) and (\textit{Frown}, \textit{Purple}), but not (\textit{Smile}, \textit{Purple}) and (\textit{Frown}, \textit{Yellow}) due to the limited number of such samples, which deteriorates the model's performance. Thus, we need more samples with rich diversity (\textit{e.g.}, (\textit{Smile}, \textit{Purple}) and (\textit{Frown}, \textit{Yellow})). To this end, we swap the latent vectors to synthesize more diverse samples.

By three pairs of robust and biased extractors for three modalities, we get robust and biased latent vectors of each modality (\ie $\textbf{v}_R^m$ and $\textbf{v}_B^m$). To diversify samples, we utilize these preliminary disentangled features for swap. We propose to generate diverse samples in latent embedding space by swapping biased latent vectors. More specifically, we replace each biased vector $\textbf{v}_{B}^m$ with a randomly selected biased vector $\hat{\textbf{v}}_{B}^m\in \mathbb{R}^{d_{s}}$ in the same mini-batch. 

To synthesize diverse samples, we concatenate robust and corresponding biased latent vectors, and also concatenate robust and randomly selected biased latent vectors. In detail, concatenated vectors $\textbf{v}^m$ are as follows,
\begin{equation}
    \textbf{v}^m = [\textbf{v}^m_R;\textbf{v}^m_B] ,
\end{equation}
where $\textbf{v}^m\in \mathbb{R}^{2d_{s}}$ denote the latent vectors that are combined with the robust latent vectors and biased latent vectors without swapping. Then, we concat $\textbf{v}_{R}^m$ and $\hat{\textbf{v}}_{B}^m$ to obtain $\hat{\textbf{v}}^m$ as follows,
\begin{equation}
    \hat{\textbf{v}}^m = [\textbf{v}_{R}^m;\hat{\textbf{v}}_{B}^m], 
\end{equation}
where $\hat{\textbf{v}}^m\in \mathbb{R}^{2d_{s}}$ represent the latent vectors that are combined with robust latent vectors and swapped biased latent vectors. Thus, by swapping, we acquire additional latent vectors $\hat{\textbf{v}}^m$ that have the same robust latent vector but a different biased latent vector with $\textbf{v}^m$. By this, we can get more samples with diverse (robust features, biased features) combinations. 

 To make the disentangled representation learning module meet more diverse samples and gain a stronger ability of disentanglement, $\textbf{v}^m$ and $\hat{\textbf{v}}^m$ are both fed into pairs of robust and biased linear layers $(L_R^m, L_B^m)$, which extract robust features $\textbf{f}_R^m,\hat{\textbf{f}}_R^m \in \mathbb{R}^{d_{s}} $ and biased features $\textbf{f}_B^m,\hat{\textbf{f}}_B^m \in \mathbb{R}^{d_{s}} $ of each modality as follows,
 \begin{equation}
 \label{eq:robust_feature}
\left\{
\begin{array}{l}  
\textbf{f}_{\kappa}^m = L_{\kappa}^m(\textbf{v}^m) = ReLU(\textbf{W}^m_{{\kappa}}\textbf{v}^m+\textbf{b}_{{\kappa}}^m), \\[5pt]
\hat{\textbf{f}}_{\kappa}^m = L_{\kappa}^m(\hat{\textbf{v}}^m) = ReLU(\textbf{W}^m_{{\kappa}}\hat{\textbf{v}}^m+\textbf{b}_{{\kappa}}^m), 
\end{array} 
\right.
\end{equation}
where $\textbf{W}^m_{{\kappa}}\in \mathbb{R}^{(d_{s})\times 2d_{s}}$, 
$\textbf{b}_{{\kappa}}^m \in \mathbb{R}^{d_{s}}$, and $ReLU(\cdot)$ is the relu activation function~\cite{nair2010rectified}. 


 \subsection{Bias Estimation}
To estimate the bias precisely, we need high-quality bias features. Thus, bias estimation has two steps, 1) training biased extractors to acquire high-quality bias features and 2) utilizing bias features to estimate bias in each modality and calculate bias weight of samples. 

\subsubsection{GMAE loss}
To facilitate biased extractors to gain high-quality bias features, we develop GMAE loss. 
It is known that the biased features are easier to learn than the robust features in the early stage of training~\cite{nam2020learning}. Based on this observation, prior studies employ GCE loss to train a biased model by amplifying the learning of ``easier'' bias. To be specific, GCE loss can make the biased model emphasize the ``easier'' samples with strong agreements between the predictions of the biased model and the labels, which amplifies the ``prejudice'' of the biased model. This is because the ``easier'' samples in the early training stage are more likely to be biased and hence the model makes more accurate predictions for biased samples.
However, GCE loss is elaborately designed for classification tasks and cannot be employed for a regression task such as MSA. Thus, we develop GMAE loss to amplify the prejudice specifically for the regression task. 
The designed GMAE loss is as follows,
 \begin{equation}
\left\{
\begin{array}{l}  
Y^m_{B} = \textbf{w}^m_{B}\textbf{f}_{B}^m+ b^m_{B},  \\[5pt]
\mathcal{L}_{GMAE}^m(Y, Y^m_{B}) = -2\ln(e^{\lvert Y-Y^m_{B} \rvert}+1) + 2\lvert Y-Y^m_{B} \rvert,
\end{array} 
\right.
\end{equation}
where $\textbf{w}^m_{B}\in \mathbb{R}^{1\times d_s}$ and $b^m_{B}\in \mathbb{R}$ are trainable parameters. $\ln(\cdot)$ denotes natural logarithm, and $\lvert\ \cdot \rvert$ denotes absolute value. To calculate GMAE loss, we forward the biased features for prediction, $Y^m_{B}\in \mathbb{R}$ are the sentiment predictions based on biased features $\textbf{f}_{B}^m$ without swapping process.

The gradient of the GMAE loss up-weights the gradient of MAE loss when the sample has a low absolute value between the prediction and the label as follows,
\begin{equation}
    \nabla\mathcal{L}_{GMAE}(Y, Y^m_{B}) = \frac{2}{1+e^{\lvert Y-Y^m_{B} \rvert}}\nabla \mathcal{L}_{MAE}(Y, Y^m_{B}).
\end{equation}
We assign greater weights to samples that are predicted well by the biased model, \textit{i.e., }the lower $\lvert Y-Y^m_{B} \rvert$, the higher $\frac{2}{1+e^{\lvert Y-Y^m_{B} \rvert}}$. In addition, GMAE loss is able to keep the gradient weight between 0 and 1, which avoids gradient explosion and makes the training process more stable. To sum up, by GMAE loss, a sample with an ``easier'' biased feature could gain low absolute value and high gradient weight while training, which helps the biased model amplify the ``prejudice''.

Meanwhile, to make biased models learn biased features from swapped samples, we utilize swapped labels $\hat{Y}^m$ for calculating GMAE loss as follows,
 \begin{equation}
\left\{
\begin{array}{l}  
\hat{Y}^m_{B} = \textbf{w}^m_{B}\hat{\textbf{f}}_{B}^m + b^m_{B},  \\[5pt]
\hat{\mathcal{L}}_{GMAE}^m(\hat{Y}^m, \hat{Y}^m_{B}) = -2\ln(e^{\lvert \hat{Y}^m-\hat{Y}^m_{B} \rvert}+1) + 2 \lvert \hat{Y}^m-\hat{Y}^m_{B} \rvert,
\end{array} 
\right.
\end{equation}
where swapped labels $\hat{Y}^m$ are along with the same selected sample of $\hat{\textbf{v}}^m$ to make biased models focus on the bias information, and  $\hat{Y}^m_{B}\in \mathbb{R}$ are the sentiment predictions based on biased features $\hat{\textbf{f}}_{B}^m$ with swapping process.

\subsubsection{Bias weight Calculation}
The biased extractors are trained with amplifying the ``prejudice'' by GMAE loss so that biased models are good at utilizing biased features for prediction. The more precisely the biased model predicts, the more biased the sample is. Thus, we employ the absolute value calculated between the prediction and the label to measure how much each modality is likely to be biased. The smaller the absolute value, the larger the bias in the modality. Then, we estimate the bias weight of a sample by calculating the minimum or average value of absolute value in each modality and taking the inverse as follows,
 \begin{equation}
 \label{eq:strategy}
\left\{
\begin{array}{l}  
\psi_{min}(Y,Y^m_{B}) = \frac{1}{\min(\lvert Y-Y^t_{B} \rvert, \lvert Y-Y^a_{B}\rvert, \lvert Y-Y^v_{B}\rvert)}, \\[10pt]
\psi_{avg}(Y,Y^m_{B}) = \frac{1}{\operatorname{avg}(\lvert Y-Y^t_{B} \rvert, \lvert Y-Y^a_{B}\rvert, \lvert Y-Y^v_{B}\rvert)},
\end{array} 
\right.
\end{equation}
where $\psi(\cdot)$ denotes the bias weight estimation function of a sample, the larger the bias weight is, the more bias a sample has. We regard the two equations in Eqn.($\ref{eq:strategy}$) as \textit{MinStrategy} and \textit{AvgStrategy}, respectively. We consider that \textit{MinStrategy} selects the most biased modality to indicate how much a sample is biased and \textit{AvgStrategy} estimates the bias degree of a sample based on the biased degree of the three modalities simultaneously. 

\subsection{Debiasing Optimization}
In this module, we first fuse multimodal robust features by multi-head self-attention. To learn robust representations from biased data with spurious correlations, we use IPW-enhanced MAE loss for training, where a sample with strong bias will be assigned with a small weight for training. Finally, we calculate the overall training objective for debiasing optimization.

\subsubsection{Robust Features Fusion}
Due to the superior performance of sentiment analysis brought by multimodal features, we develop a multimodal fusion mechanism for final prediction. And to reinforce the generalization ability of the model, we utilize the robust features for fusion.

First, we stack the three robust features (from Eqn.(\ref{eq:robust_feature})) into a matrix $\textbf{M} = [\textbf{f}_R^t, \textbf{f}_R^a, \textbf{f}_R^v] \in \mathbb{R}^{3\times d_{s}}$. Then, in order to make each vector aware of its companion cross-modal features, we employ a multi-head self-attention on these features. By doing this, each feature is given the opportunity to gain consistent and complementary information from other features that could contribute to the overall sentiment analysis. Specifically, suppose we have $U$ attention heads, and the attention function of the $i$-th attention head can be formulated as follows,
\begin{equation}
\label{eq:transformer_head}
\left\{
\begin{array}{l}  
\textbf{Q}_i  = \textbf{M}\textbf{W}^q_i, 
\textbf{K}_i  = \textbf{M}\textbf{W}^k_i, 
\textbf{V}_i  = \textbf{M}\textbf{W}^v_i, \\
\textbf{O}_i  = softmax(\textbf{Q}_i\textbf{K}_i^T/\sqrt{d_{s}})\textbf{V}_i,
\end{array} 
\right.
\end{equation}
where $\textbf{Q}_i, \textbf{K}_i, \textbf{V}_i\, \textbf{O}_i\in \mathbb{R}^{3\times \frac{d_{s}}{U}}$ are the query, the key, and the value projected from the matrix $\textbf{M}$, respectively. $\textbf{W}_i^{q/k/v}\in \mathbb{R}^{d_{s}\times \frac{d_{s}}{U}}$ are learnable matrices in the $i$-th attention head. The multi-head self-attention outputs a matrix $\hat{\textbf{M}}\in \mathbb{R}^{3\times d_{s}}$ as follows,
\begin{equation}
    \hat{\textbf{M}} = [\textbf{O}_1;\dots;\textbf{O}_U]\textbf{W}^o,
\end{equation}
where $\textbf{W}^o\in \mathbb{R}^{d_{s}\times d_{s}}$, and each $\textbf{O}_i$ here is calculated based on Eqn.(\ref{eq:transformer_head}). Finally, we take the multi-head self-attention output $\hat{\textbf{M}}\in \mathbb{R}^{3\times d_{s}}$ and construct a joint-vector $\textbf{f}_{o} \in \mathbb{R}^{3d_{s}}$ using concatenation. The final sentiment predictions are then generated by a classifier as follows,
\begin{equation}
    \tilde{Y} = \textbf{w}\textbf{f}_o + b,
\end{equation}
where $\textbf{w}\in \mathbb{R}^{3d_{s}}$ and $b\in \mathbb{R}$. 
Meanwhile, we also fuse the robust features $\hat{\textbf{f}}_R^m$ disentangled from swapped concatenated vectors in the same way as above to calculate $\hat{Y}$.

\subsubsection{IPW-enhanced MAE loss}
For the regression task, existing studies mostly utilize MAE loss as follows,
\begin{equation}
    \mathcal{L}_{MAE} = \lvert Y-\tilde{Y} \rvert .
\end{equation}
However, MAE loss treats samples with/without bias equally. Training with MAE loss, robust extractors cannot focus on samples without bias to learn robust features, which acquire features that contain biased features from biased samples. We thus can train robust extractors by making them focus on learning unbiased samples. Our robust extractors are unable to extract robust features from unbiased samples since they cannot recognize which sample is unbiased. To learn robust features from biased data with spurious correlations, a widely used method is IPW~\cite{qi2022class}, where a sample with strong bias will be assigned with a small weight for training. This helps robust extractors focus on learning robust features of unbiased samples. Thus, we utilize IPW-enhanced MAE loss for training, which re-weight the MAE loss by bias weight as follows,
\begin{equation}
    \mathcal{L}_{IPW} = \mathcal{L}_{MAE}\cdot \frac{1}{P(x|Bias(x))+1}.
    \label{eq:ipw1}
\end{equation}
This IPW implies that if a sample $x=[T, A, V]$ is more likely associated with its biased features (\textit{i.e., }$Bias(x)$), we should underweight the loss to discourage such a biased sample. We calculate $P(x|Bias(x))$ as follows,
\begin{equation}
    P(x|Bias(x)) \propto \psi(Y,Y^m_{B}) \cdot (\lvert Y-\tilde{Y} \rvert),
\end{equation}
where $\psi(\cdot)$ is illustrated in Eqn.(\ref{eq:strategy}), and $\lvert Y-\tilde{Y} \rvert$ is the absolute value without gradients.

Meanwhile, we also calculate IPW-enhanced MAE loss for the swapped sample. To make robust extractors focus on robust information, we employ $Y$ as the label of the swapped sample and the same weight for MAE loss as follows,
 \begin{equation}
\left\{
\begin{array}{l}  
\hat{\mathcal{L}}_{MAE} = \lvert Y-\hat{Y} \rvert ,  \\[5pt]
\hat{\mathcal{L}}_{IPW} = \hat{\mathcal{L}}_{MAE} \cdot \frac{1}{P(x|Bias(x))+1}.
\end{array} 
\right.
\label{eq:ipw2}
\end{equation}

\subsubsection{Training Objective}
The overall learning objective of the model is performed by minimizing,
\begin{equation}
    \mathcal{L} = \mathcal{L}_{IPW} +  \lambda\mathcal{L}_{GMAE}^m +  \beta(\hat{\mathcal{L}}_{IPW} +\lambda\hat{\mathcal{L}}_{GMAE}^m) ,
    \label{eq:overall}
\end{equation}  
where $\lambda$ and $\beta$ are adjusted for weighting the importance of GMAE loss and swap, respectively. To ensure that robust modules and biased modules focus on robust attributes and biased attributes, respectively, $\mathcal{L}_{IPW}$ and $\hat{\mathcal{L}}_{IPW}$ are backpropagated to robust features fusion, robust linear, and robust extractor, $\mathcal{L}_{GMAE}^m$ and $\hat{\mathcal{L}}_{GMAE}^m$ are backpropagated to linear, biased linear, and biased extractor.

\section{Experiments}
In this section, we conducted extensive experiments on two widely-used benchmark datasets, (\ie MOSI and MOSEI), to answer the following research questions.
\begin{itemize}[leftmargin=*]
  \item \textbf{RQ 1:} Does GEAR outperform state-of-the-art MSA baselines on the OOD testing sets?
  \item \textbf{RQ 2:} How does GEAR perform on the IID testing set?
  \item \textbf{RQ 3:} How does each component affect GEAR?
  \item \textbf{RQ 4:} How is the qualitative performance of GEAR?
\end{itemize}

\subsection{Experimental Settings}

\subsubsection{Datasets}




\begin{table*}[t]
\setlength{\abovecaptionskip}{0.2cm}
\setlength{\belowcaptionskip}{0cm}
\centering
\caption{IID and OOD testing performance (\%) comparison among different methods on MOSEI datasets. For \textit{Acc-2} and \textit{F1}, we reported results on both these metrics using the segmentation marker $\text{-}/\text{-}$ where the left-side score is for $neg./nonneg.$ while the right-side score is for $neg./pos$. The AVG (OOD) means the average result over four OOD sets. $Imp.$ denotes the improvement of our model compared to the best-performing baseline. The best result is highlighted in bold and the second-best result is underlined. ${}^{\dagger} p<0.05$ under McNemar’s Test for accuracy improvement compared with all baselines.}
\setlength{\tabcolsep}{0.5mm}
\resizebox{\textwidth}{!}{
\begin{tabular}{c|cc|cc|cc|cc|cc|cc}
\toprule 
\multicolumn{1}{c|}{\multirow{3}{*}{Model}} & \multicolumn{12}{c}{MOSEI} \\
& \multicolumn{2}{c}{IID} & \multicolumn{2}{c}{OOD Text} & \multicolumn{2}{c}{OOD Audio}& \multicolumn{2}{c}{OOD Video} & \multicolumn{2}{c}{OOD TAV} & \multicolumn{2}{c}{AVG (OOD)}  \\ \cline{2-13} 
& \textit{Acc} & \textit{F1} & \textit{Acc} & \textit{F1} & \textit{Acc} & \textit{F1}& \textit{Acc} & \textit{F1}& \textit{Acc} & \textit{F1}& \textit{Acc} & \textit{F1}  \\ \hline
MISA     &82.91/\underline{85.47}&83.11/\underline{85.28}&79.06/\underline{81.23}&79.00/81.06&80.09/81.73&80.03/81.89&80.64/\underline{81.70}&80.57/\underline{81.86}&77.32/\underline{79.60}&77.26/\underline{79.80}&79.28/\underline{81.07}&79.22/\underline{81.15}  \\
MAG-BERT &83.14/84.61&83.27/84.40&79.15/80.06&79.00/79.86&80.40/81.07&80.32/81.23&80.65/80.84&80.55/80.99&77.83/78.58&77.73/78.79&79.51/80.14&79.40/80.22  \\
Self-MM  &\textbf{84.69}/85.35&\textbf{84.69}/85.10&80.53/80.60&80.31/80.39&\underline{81.53}/81.47&\underline{81.42}/81.63&81.40/81.01&81.28/81.15&\underline{78.75}/78.65&78.62/78.85&\underline{80.55}/80.43&80.41/80.51  \\
CENet    &82.77/85.13&83.00/84.97&78.61/80.73&78.56/80.57&80.32/\underline{81.81}&80.26/\underline{81.96}&80.89/\underline{81.70}&80.81/81.85&77.29/79.28&77.22/79.48&79.28/80.88&79.21/80.97  \\
CubeMLP  &83.38/85.05&83.48/84.81&79.33/80.44&79.16/80.22&80.65/81.48&80.56/81.63&81.05/81.37&80.95/81.52&77.98/79.00&77.88/79.19&79.75/80.57&79.64/80.64  \\
CLUE     &83.99/85.06&83.90/85.26&\underline{80.91}/81.09&\textbf{81.14}/\underline{81.33}&81.03/80.72&81.16/80.57&\underline{81.54}/80.95&\underline{81.70}/80.82&78.48/78.48&\underline{78.65}/78.06&80.49/80.31&\underline{80.66}/80.20  \\
\hline
GEAR$^{\dagger}$     &\underline{84.06}/\textbf{85.88}&\underline{84.30}/\textbf{85.79}&\textbf{80.99}/\textbf{82.33}&\underline{80.97}/\textbf{82.24}&\textbf{82.45}/\textbf{83.48}&\textbf{82.42}/\textbf{83.64}&\textbf{82.51}/\textbf{83.05}&\textbf{82.48}/\textbf{83.21}&\textbf{79.85}/\textbf{81.22}&\textbf{79.82}/\textbf{81.41}&\textbf{81.45}/\textbf{82.52}&\textbf{81.42}/\textbf{82.63}  \\
\bottomrule	
\end{tabular}
}
\label{tab:mosei}
\end{table*}

\begin{table*}[t]
\setlength{\abovecaptionskip}{0.2cm}
\setlength{\belowcaptionskip}{0cm}
\centering
\caption{IID and OOD testing performance (\%) comparison among different methods on MOSI datasets. The explanations of notations are the same as those in Table~\ref{tab:mosei}. }
\setlength{\tabcolsep}{0.5mm}
\resizebox{\textwidth}{!}{
\begin{tabular}{c|cc|cc|cc|cc|cc|cc}
\toprule 
\multicolumn{1}{c|}{\multirow{3}{*}{Model}} & \multicolumn{12}{c}{MOSI} \\
& \multicolumn{2}{c}{IID} & \multicolumn{2}{c}{OOD Text} & \multicolumn{2}{c}{OOD Audio}& \multicolumn{2}{c}{OOD Video} & \multicolumn{2}{c}{OOD TAV} & \multicolumn{2}{c}{AVG (OOD)}  \\ \cline{2-13} 
& \textit{Acc} & \textit{F1} & \textit{Acc} & \textit{F1} & \textit{Acc} & \textit{F1}& \textit{Acc} & \textit{F1}& \textit{Acc} & \textit{F1}& \textit{Acc} & \textit{F1}  \\ \hline
MISA     &82.66/84.2&82.65/84.24&78.42/79.53&78.41/79.54&81.24/83.14&81.22/83.13&81.33/82.91&81.31/82.91&76.97/79.19&76.95/79.20&79.49/81.19&79.47/81.20  \\
MAG-BERT &\underline{83.04}/\underline{84.81}&\underline{83.00}/\underline{84.82}&79.20/80.76&79.18/80.76&\textbf{82.29}/\textbf{84.37}&\textbf{82.24}/\textbf{84.35}&81.59/83.62&81.57/83.61&77.55/79.92&77.51/79.92&\underline{80.16}/82.17&\underline{80.13}/82.16  \\
Self-MM  &82.95/\underline{84.81}&82.87/84.79&79.20/81.17&79.17/81.17&81.36/83.66&81.29/83.61&\underline{81.66}/\underline{83.76}&\underline{81.63}/\underline{83.74}&\underline{78.23}/\underline{80.54}&\underline{78.18}/\underline{80.52}&80.11/\underline{82.28}&80.07/\underline{82.26}  \\
CENet    &82.27/84.09&82.14/84.03&\underline{79.79}/\underline{81.28}&\underline{79.71}/\underline{81.22}&80.68/82.88&80.58/82.80&81.19/83.26&81.12/83.22&78.04/80.33&77.97/80.30&79.93/81.94&79.85/81.89  \\
CubeMLP  &81.73/83.64&81.60/83.58&79.40/\underline{81.28}&79.30/81.21&79.94/82.10&79.85/82.04&80.52/82.70&80.45/82.65&\underline{78.23}/80.33&78.14/80.27&79.52/81.60&79.44/81.54  \\
CLUE     &82.79/84.68&82.87/84.71&78.97/81.07&79.03/81.09&80.56/82.80&80.60/82.83&81.32/83.55&81.35/83.56&77.38/79.79&77.43/79.81&79.56/81.80&79.60/81.82  \\
\hline
GEAR$^{\dagger}$     &\textbf{83.29}/\textbf{84.96}&\textbf{83.22}/\textbf{84.95}&\textbf{80.47}/\textbf{82.10}&\textbf{80.44}/\textbf{82.09}&\underline{82.22}/\underline{84.31}&\underline{82.16}/\underline{84.27}&\textbf{82.53}/\textbf{84.39}&\textbf{82.50}/\textbf{84.37}&\textbf{79.98}/\textbf{82.09}&\textbf{79.94}/\textbf{82.08}&\textbf{81.30}/\textbf{83.22}&\textbf{81.26}/\textbf{83.20}  \\
\bottomrule	
\end{tabular}
}
\label{tab:mosi}
\end{table*}

To demonstrate the effectiveness of our GEAR, we conducted extensive experiments on MOSI and MOSEI datasets, which are widely used in the MSA task.
\begin{itemize}[leftmargin=*]
  \item \textbf{MOSI}~\cite{zadeh2016mosi} is a publicly released  MSA dataset. It collects 2,199 utterance-video clips of 93 monologue videos from YouTube platform\footnote{https://www.youtube.com.}, each of which is labeled with a continuous sentiment score ranging from -3 (strongly negative) to 3 (strongly positive). 
  
  \item \textbf{MOSEI}~\cite{zadeh2018multimodal} is an expanded version of MOSI. In MOSEI, 3,837 monologue videos are also collected from YouTube, involving 250 topics and 22,856 utterance-level labeled instances, each of which is also labeled with a continuous sentiment score ranging from -3 to 3.  
\end{itemize}
The video clips in the two datasets consist of textual descriptions, acoustic tracks, and visual keyframes, which provide multimodal information to reflect the sentiment. 



\subsubsection{IID and OOD Settings} 


We removed the spurious correlations by discarding samples from the IID testing set to build the OOD testing sets. 
Due to the different bias types across modalities, we employed different strategies to construct the OOD datasets for different modalities. 
For the OOD Text set, following Sun ~\textit{et al.}~\cite{sun2022counterfactual}, we adopt the same method as described in this paper and refer to it for further details. We first obtained the distribution of word frequency in different sentiment categories. Then, we used the simulated annealing algorithm for dataset construction, which iteratively optimizes the OOD Text set to make the distribution of all words on different sentiment categories as same as possible, (\textit{e.g., }the word ``movie'' has an equal number of positive and negative categories). 
For the OOD Audio set and OOD Video set, the attributes of the audio and video are not recognizable by humans, and thus the distributions of each attribute are inaccessible. To mitigate this issue, we employ K-means clustering on hand-crafted features provided by \cite{yu2021learning} of the audio or video to obtain $k$ clusters for each modality ($k=100$ for audio and video). We assume that each cluster derived from K-means represents an attribute, and for the two OOD datasets, we ensure that all attributes appear equally in different sentiment categories by random sampling. For example, if there are four positive samples and six negative samples in a cluster, we randomly sample four samples from the six negative samples to ensure that the number of samples in each category is the same, (\textit{e.g., }make ``blue background'' appear equally in positive and negative categories). 
For OOD TAV set, we first obtained the distribution of word frequency and attributes in different sentiment categories as mentioned above. Following the existing work~\cite{sun2022counterfactual}, we employed the simulated annealing algorithm to make the distribution of all words and all attributes on different sentiment categories as same as possible simultaneously. 

\vspace{-1pt}
\subsubsection{Evaluation Tasks and Metric}
Due to space limitations and following the latest work~\cite{sun2022counterfactual}, we mainly focus on the metrics of accuracy and F1 score.
Two distinct formulations have been considered in the past. The first is $negative/non\text{-}negative$ where $negative$ denotes a class with sentiment scores $<0$ and $non\text{-}negative$ class with sentiment scores $>=0$. Second, recent work~\cite{hazarika2020misa} employs a more accurate formulation of $negative/positive$ classes where negative and positive classes are assigned with $<0$ and $>0$ sentiment scores, respectively. For the fair competition, we reported both $negative/non\text{-}negative$ and $negative/positive$ results. We converted the predicted score on the regression task into these two formulations, and then we used Accuracy and Weighted-F1 to measure the performance of the models.

\vspace{-1pt}
\subsubsection{Baselines}
To evaluate the performance of GEAR, we employed the following methods for comparison.
\begin{itemize}[leftmargin=*]

  \item \textbf{MISA~\cite{hazarika2020misa}:} The model projects modalities into model-specific and model-invariant vectors, capturing cross-modal interactions.
  \item \textbf{MAG-BERT~\cite{rahman2020integrating}:} This baseline employs the nonverbal representations with sentimental polarity to shift lexical representations within the pretrained language model.
  \item \textbf{Self-MM~\cite{yu2021learning}:} This model develops a unimodal sentiment label-generating module based on a self-supervised method to aid in learning modality-specific representations. 
  \item \textbf{CENet~\cite{wang2022cross}:} This baseline employs an attention-based gate to capture asynchronous emotion cues from unaligned data.
  \item \textbf{Cube-MLP~\cite{sun2022cubemlp}:} This baseline develops MLPs to mix features on three dimensions: sequence, modality, and channel.
  \item \textbf{CLUE~\cite{sun2022counterfactual}:} This framework captures the direct effect of textual modality via an extra text model and estimates the total effect by an MSA model. 
  
\end{itemize}





\subsubsection{Implementation Details}

We implemented all baselines and our GEAR using Pytorch\footnote{\url{https://pytorch.org}.}. To optimize the parameters of the models, we adopted Adam~\cite{kingma2014adam} optimizer with a learning rate of $5e\text{-}5$ for BERT and $1e\text{-}3$ for other modules. Note that, as we needed well-disentangled robust and biased features for swap, we began to swap after certain epochs $e_s$.  We employed a grid search strategy to identify the optimal hyperparameters for our model. Specifically, we set the batch size $N$ to 32, the latent vector dimension $d_s$ to 32, the head number $U$ to 4, and the swap weight $\beta$ to 0.3. Additionally, we set the swap epochs $e_s$ to 8 and 11, and the GMAE weight $\lambda$ to 10 and 18, for MOSEI and MOSI, respectively. Besides, we employed the early stopping strategy, which stops the training if the accuracy score/loss does not increase/decrease for 8 successive epochs. 
For all baselines, we used the grid search strategy to find the optimal parameter settings to achieve the best performance. 
For a fair comparison, we reported the average experimental results on accuracy and F1 score over three random seeds.



\subsection{Model Comparison (RQ1 \& RQ2)}
We conducted experiments on the IID and OOD testing sets of MOSEI and MOSI datasets, respectively. As shown in Tables~\ref{tab:mosei} and~\ref{tab:mosi}, we had the following observations. 
1) The average accuracy on $neg./pos$ was observed to increase by 1.46\% on the MOSEI dataset and by 1.14\% on the MOSI dataset. GEAR achieves clear margins over the prior methods on OOD sets, which demonstrates that GEAR has superior debiasing ability over existing methods. After conducting significant tests, $p<0.05$ proves that our results are significant.
2) In particular, the improvements are most obvious in the OOD TAV testing set. The possible reason is that the OOD TAV testing set removes spurious correlations in all three modalities and the models' general debiasing ability removed bias in three modalities. 
3) The improvement of GEAR on the IID testing set is smaller than on OOD testing sets, which indicates that the bias between the training set and the IID testing set is very small and GEAR's debiasing ability cannot be fully utilized.
4) On the unimodal and multimodal OOD testing sets, all methods perform worse than on IID testing sets. This demonstrates that methods indeed suffer from the spurious correlation for prediction. In spite of this, the performance decrease of GEAR is the least compared with that of baseline methods. 
And 5) CLUE performs well on the OOD Text set of the MOSEI dataset, for which we reasoned that CLUE is specifically designed for reducing spurious correlations between textual words and sentiment labels. However, on the other two unimodal testing sets and the multimodal testing set, CLUE performs worse than our GEAR. One reasonable explanation is that CLUE has limited ability in debiasing the acoustic and visual modalities. 

\begin{table}[t]
\setlength{\abovecaptionskip}{0.2cm}
\setlength{\belowcaptionskip}{0cm}
\centering
\caption{Ablation study results (\%) for $neg./pos.$ results of GEAR on OOD TAV testing sets of MOSEI and MOSI dataset.
The best results are highlighted in boldface. 
}
\setlength{\tabcolsep}{5mm}{
\resizebox{0.48\textwidth}{!}{
\begin{tabular}{l|cc|cc}
\toprule 
\multicolumn{1}{c|}{\multirow{2}{*}{Model}}    
& \multicolumn{2}{c|}{MOSEI} & \multicolumn{2}{c}{MOSI}\\ 
& \textit{Acc} & \textit{F1} & \textit{Acc} & \textit{F1} \\ \hline
GEAR (OOD TAV)  &\textbf{81.22}&\textbf{81.41}&\textbf{82.09}&\textbf{82.08} \\
$\quad$w/o-IPW    &78.84&79.03&81.37&81.29         \\ 
$\quad$w/o-GMAE   &79.88&80.08&80.12&80.12    \\ 
$\quad$w/o-Swap   &80.32&80.51&81.47&81.49         \\ 
\hline
GEAR (OOD Text) &\textbf{82.33}&\textbf{82.24}&\textbf{82.10}&\textbf{82.09} \\
$\quad$w/o-Text    &81.22&81.06&81.07&81.05         \\ 
\hline
GEAR (OOD Audio) &\textbf{83.48}&\textbf{83.64}&\textbf{84.31}&\textbf{84.27} \\
$\quad$w/o-Audio   &81.75&81.91&81.77&81.78    \\ 
\hline
GEAR (OOD Video) &\textbf{83.05}&\textbf{83.21}&\textbf{84.39}&\textbf{84.37}  \\
$\quad$w/o-Video   &82.06&82.21&84.04&84.02         \\ 
\bottomrule
\end{tabular}}}
\label{tab:ablation}
\end{table}

\subsection{Ablation Study (RQ3)}
To verify the effectiveness of the main components in the proposed model, we conducted extensive ablation studies on OOD TAV, OOD Text, OOD Audio, and OOD Video datasets. We introduced several variants for analysis. 
(1) \textbf{w/o-IPW}. In this variant, we replaced $\mathcal{L}_{IPW}$ and $\hat{\mathcal{L}}_{IPW}$ with $\mathcal{L}_{MAE}$ and $\hat{\mathcal{L}}_{MAE}$ by removing the weight of MAE loss in Eq.(\ref{eq:ipw1}) and Eq.(\ref{eq:ipw2}). (2) \textbf{w/o-GMAE}. To verify the effect of the proposed GMAE loss, we trained the biased model with standard MAE loss instead of our proposed GMAE loss by replacing GMAE loss in Eq.(\ref{eq:overall}) with MAE loss. (3) \textbf{w/o-Swap}. We set swap epochs $e_s$ to an extremely large number to remove the swap operation. (4) \textbf{w/o-Text, w/o-Audio, and w/o-Video}. We removed the bias estimation of text, audio, or video modality, respectively. In other words, we calculated the bias weight only using two modalities in Eq.(\ref{eq:strategy}). These variants are tested on their corresponding OOD testing sets.

Table~\ref{tab:ablation} shows the results of the ablation studies. 
First, after employing the model w/o-IPW, we can see the performance drops significantly. This phenomenon shows assigning a small weight to a sample with a strong bias for debiasing is indeed indispensable. Second, the setting of w/o-GMAE obtains worse results than the original model. This is because our proposed GMAE loss can train a model to be biased by amplifying the prejudice. However, the model trained with standard MAE loss not only exploits the biased attribute but also partially learns the robust attribute, which can hurt the debiasing ability of our overall algorithm by estimating the bias of each modality inaccurately. Third, GEAR w/o-Swap performs marginally better than almost all models in MOSEI and MOSI datasets, but not as well as GEAR. Compared to GEAR w/o-Swap, GEAR gains the relative improvements of 0.90\% and 0.62\% evaluated by $Acc$ for the two datasets, respectively. This shows that the diversity of samples for disentanglement is crucial for optimal performance. 
Furthermore, 
we observed that after removing the bias estimation of any modality, the performance on the corresponding dataset drops significantly, which confirms that GEAR has superior general debiasing ability with each modality.

\begin{figure}
\setlength{\abovecaptionskip}{0.1cm}
\setlength{\belowcaptionskip}{0cm}
    \centering
    \includegraphics[scale=0.24]{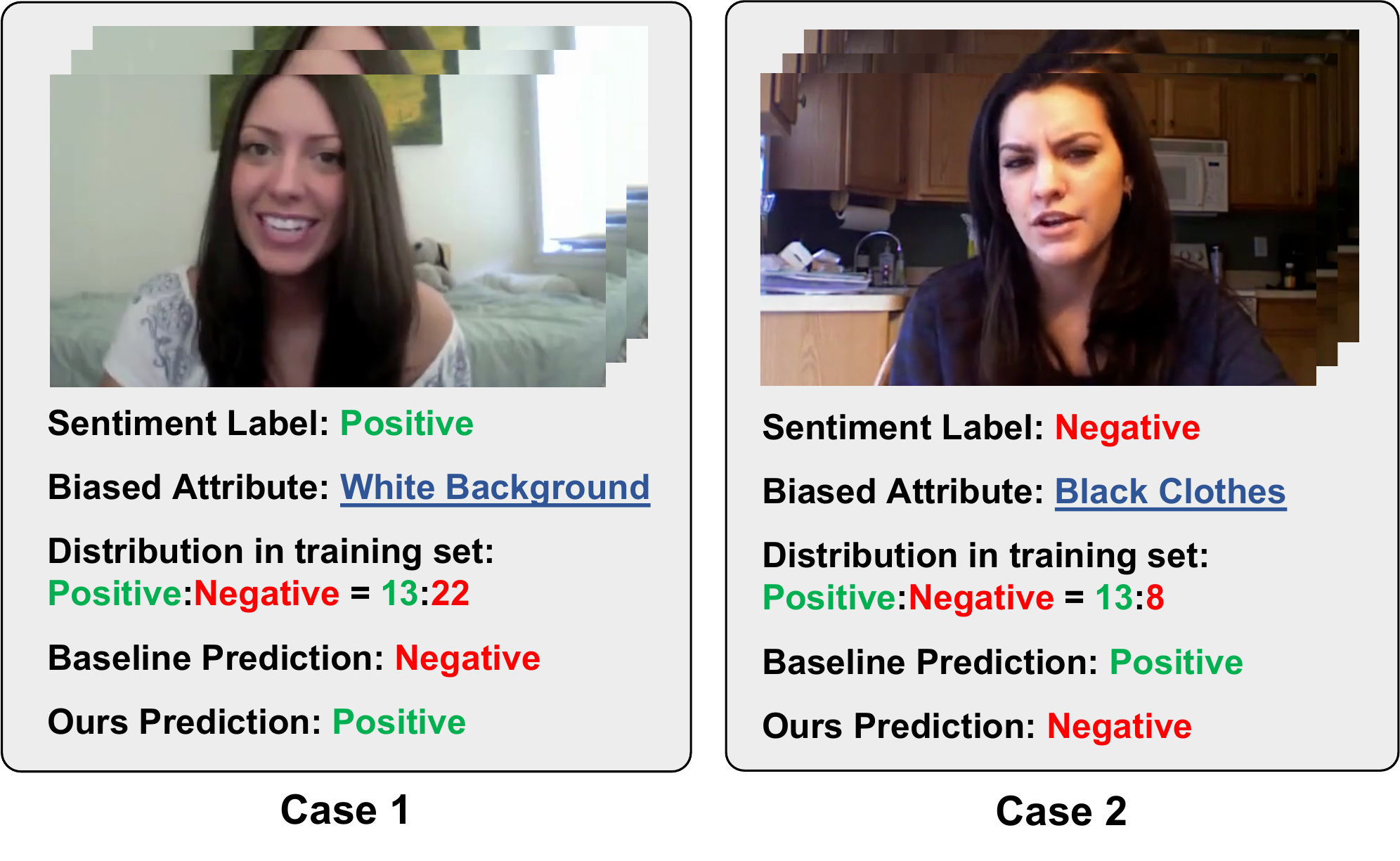}
    \caption{Two testing cases of the baseline (\ie Self-MM) and GEAR on the $neg./pos.$ results.}
    \label{fig:case4.9}
\vspace{-1em}
\end{figure}

\subsection{Case Study (RQ4)}
To gain more insights into our model, we randomly selected four cases to explain how the spurious correlations in video modality affect the traditional MSA model and why GEAR is able to handle such spurious correlations in the testing set. We illustrated the binary ($neg./pos.$) results of Self-MM and GEAR on four testing samples from MOSI datasets in Figure~\ref{fig:case4.9} because the Self-MM shows the best overall performance (\textit{i.e.,} AVG (OOD)). 
To learn the debiasing ability of our model, for each case, we recognized its biased attribute and counted the sentiment frequency of this biased attribute in the training dataset. There are 93 videos in the MOSI dataset, each of which has several clips. The clips in the same video have the same biased attributes, and thus, for convenience, our biased attributes statistics was at the video level. However, the dataset is labeled at the clip level. To obtain the video-level label, we selected the dominant label of all video clips as the video label. In detail, for a video, if there are more  positive video clips than  negative video clips, then the video is considered positive.

Taking Case 1 as an example, we can see that there is a woman with a smile and white background. By manual recognition, we found a total of 35 videos with white backgrounds from the training set, 13 with positive sentiment labels, and 22 with negative sentiment labels. And the white background is a kind of superficial feature that can be captured easily by models. Thus, the white background is a biased attribute that induces spurious correlations with sentiment labels. The traditional MSA model cannot reduce the spurious correlations, it thus predicts Case 1 as negative sentiment. Different from Self-MM, GEAR is able to handle this biased case by employing robust features such as the smile of the women for prediction. This demonstrates that GEAR has strong debiasing ability. A similar observation can be found in Case 2.


\section{Conclusion and future work}
In this work, we first point out the spurious correlations between multimodal input data and sentiment labels and formulate a general debiasing multimodal sentiment analysis task. We design a novel general debiasing framework for multimodal sentiment analysis, GEAR for short, which strengthens the generalization ability via disentangling the robust features and bias features of textual, acoustic, and visual modalities,  estimating the bias weight, and training with IPW-enhanced loss.  
Extensive experiments on two datasets (\textit{i.e.,} MOSEI and MOSI) confirm the existence of spurious correlations and also indicate the superior generalization ability of GEAR on OOD testing sets. 
In future work, we will explore new strategies such as invariant feature learning to learn better disentangle biased features and facilitate bias estimation. 

\begin{acks}
This material is in part based on research sponsored by Defense Advanced Research Projects Agency (DARPA) under agreement number HR0011-22-2-0047.
\end{acks}

\bibliographystyle{ACM-Reference-Format}
\balance
\bibliography{reference}

\end{document}